\documentclass[conference]{IEEEtran}
\IEEEoverridecommandlockouts
\usepackage{cite}

\usepackage[none]{hyphenat}
\usepackage{amsmath,amssymb,amsfonts}

\usepackage{algorithmic}
\usepackage{graphicx}
\usepackage{textcomp}
\usepackage{xcolor}
\usepackage{float}
\usepackage{booktabs}
\usepackage{array}
\usepackage{multirow}
\usepackage{threeparttable}

\usepackage[colorlinks=true, linkcolor=blue, citecolor=blue, urlcolor=blue]{hyperref}
\usepackage[noabbrev, capitalise]{cleveref}
\usepackage[table]{xcolor}
\usepackage{tikz}
\usetikzlibrary{shapes, positioning}
\usepackage[inkscapelatex=false]{svg}
\definecolor{lightgray}{gray}{0.8}
\definecolor{AliceBlue}{rgb}{0.94, 0.97, 1.0}

\def\BibTeX{{\rm B\kern-.05em{\sc i\kern-.025em b}\kern-.08em
    T\kern-.1667em\lower.7ex\hbox{E}\kern-.125emX}}

\begin{document}

\title{LoCal-RIO: Radar-Inertial Odometry with Loop-Closure IMU Bias Calibration\\
}

\author{%
    Ali Alridha Abdulkarim,
    Mikhail Litvinov,
    Zein Alabdeen Abdulkarim,
    Dzmitry Tsetserukou
\thanks{The authors are affiliated with the Intelligent Space Robotics Laboratory, Center for Digital Engineering, Skolkovo Institute of Science and Technology. \newline {\tt \{ali.abdulkarim, mikhail.litvinov2, z.abdulkarim.ext, d.tsetserukou\}@skoltech.ru}}
}

\maketitle

\begin{abstract}
Millimeter-wave radar enables robust perception in visually degraded environments, yet radar-inertial estimation remains prone to drift: body-frame velocity measurements do not constrain heading and position, and the gyroscope bias, which governs heading drift, is poorly observable over the short horizons of sliding-window estimators. We propose a hierarchical radar-inertial factor graph that separates estimation into a fixed-lag navigation graph, which fuses IMU preintegration, radar velocities, ZUPT, and ground-plane constraints into smooth, low-latency odometry, and a keyframe mapping graph, which combines this odometry with submap registration and loop closures. Loop closures additionally calibrate the IMU: the part of a loop residual explained by a bias error is estimated through preintegration Jacobians chained over the loop interval and enters the navigation graph as a prior on the bias alone. Since this calibration is irreversible, it uses only loop closures accepted by the mapping graph and a cycle-consistency test. Extensive evaluations demonstrate high accuracy and drift-reduced estimation at real-time speeds. \\ Code and datasets will be released upon paper acceptance.
\end{abstract}

\section{INTRODUCTION}
\label{sec:introduction}
Autonomous mobile robots require robust perception and state estimation. While Visual-Inertial Odometry (VIO) and LiDAR-Inertial Odometry (LIO) perform well in structured, clear environments, they degrade under obscurants such as fog, dust, or poor lighting. Millimeter-wave (mmWave) radar offers an alternative: its longer wavelengths penetrate visual obstructions, and it provides both geometry and Doppler velocity. However, radar-inertial odometry faces distinct algorithmic challenges. Radar point clouds are sparse, cluttered, and prone to multipath reflections, and indoor environments---caves, tunnels, corridors---introduce scale constraints, low reflectivity, and repetitive patterns that remain underexplored. Fusing IMU data with radar velocities enables high-frequency tracking, yet creates a fundamental observability issue: radar velocity measurements do not constrain heading, leaving the gyroscope bias weakly observable without sufficient excitation. Standard sliding-window estimators marginalize older states, truncating the temporal baseline needed for accurate bias estimation and causing drift. To achieve global consistency, hierarchical methods typically add a backend optimization layer. However, the backend corrections either reach the control odometry as discontinuities or remain confined to the map, while the frontend continues to integrate a biased sensor, leaving the root cause of drift unaddressed.

\begin{figure}[ht]
    \centering
    \makebox[\linewidth][c]{%
        \includegraphics[width=0.96\linewidth]{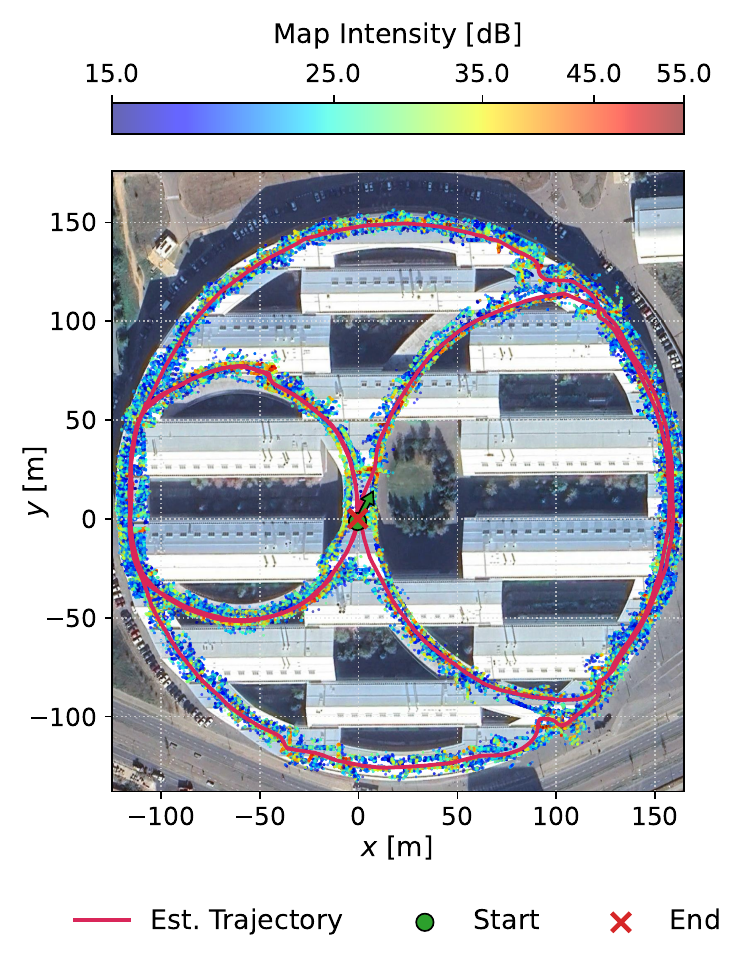}%
    }
\caption{Estimated trajectory and radar map overlaid on a satellite image of the building featured in the campus dataset. The building's three enclosed rings (perimeters: 858 m, 593 m, 376 m) were mapped over a total loop distance of 2203.5 m, with the smallest ring traversed twice. The map reflects the spatial geometry and curvature of the building, demonstrating consistent submap matching and re-localization in a large indoor environment.}
    \vspace{-0.12in}
    \label{fig:satellite_overlay}
\end{figure}

\vspace{4em}
We present LoCal-RIO, a hierarchical radar-inertial odometry framework with two factor graphs. The frontend is a fixed-lag navigation graph fusing IMU preintegration, radar velocities, Zero-Velocity Updates (ZUPT), and ground-plane constraints to produce smooth, high-rate odometry for real-time control. The backend is a mapping graph over all keyframe poses. It combines the relative motion of the navigation graph with submap registrations and loop closures, and estimates the transform between the odometry and map frames, which absorbs all loop-closure corrections. Loop closures reach the navigation graph only through the IMU bias: the drift accumulated between two visits is regressed onto the bias, and the result is added as a prior on the bias alone.

The primary contributions of this work are:
\begin{itemize}
\item \textbf{Loop Closure as Bias Calibration:} Loop residuals are regressed onto the IMU bias through chained preintegration Jacobians. The resulting bias prior corrects the source of drift without altering poses, keeping the odometry continuous and suitable for robot control.

\item \textbf{Consistency-Informed Loop Acceptance:} Loop closures are weighted incrementally with a graduated robust kernel initialized by a pairwise cycle-consistency statistic, and are used for bias calibration only after subsequent loop closures have tested them.

\item \textbf{Specular-Filtered Submapping:} A trajectory-to-map coupling mechanism that (i) transforms raw scans using finalized smoother poses at the marginalization boundary to prevent geometric submap blurring, and (ii) replaces centroid-averaged voxelization with a maximum-SNR selection criterion for the map and place recognition clouds, preserving raw coordinates of specular targets while discarding low-power multipath clutter.

\item \textbf{Empirical Validation:} Real-world evaluation demonstrating that LoCal-RIO mitigates drift and tracking failures common to radar-inertial and LiDAR-inertial SLAM systems.
\end{itemize}

\begin{figure*}[t]
    \vspace*{-1em}
    \centering
    \includegraphics[width=\textwidth]{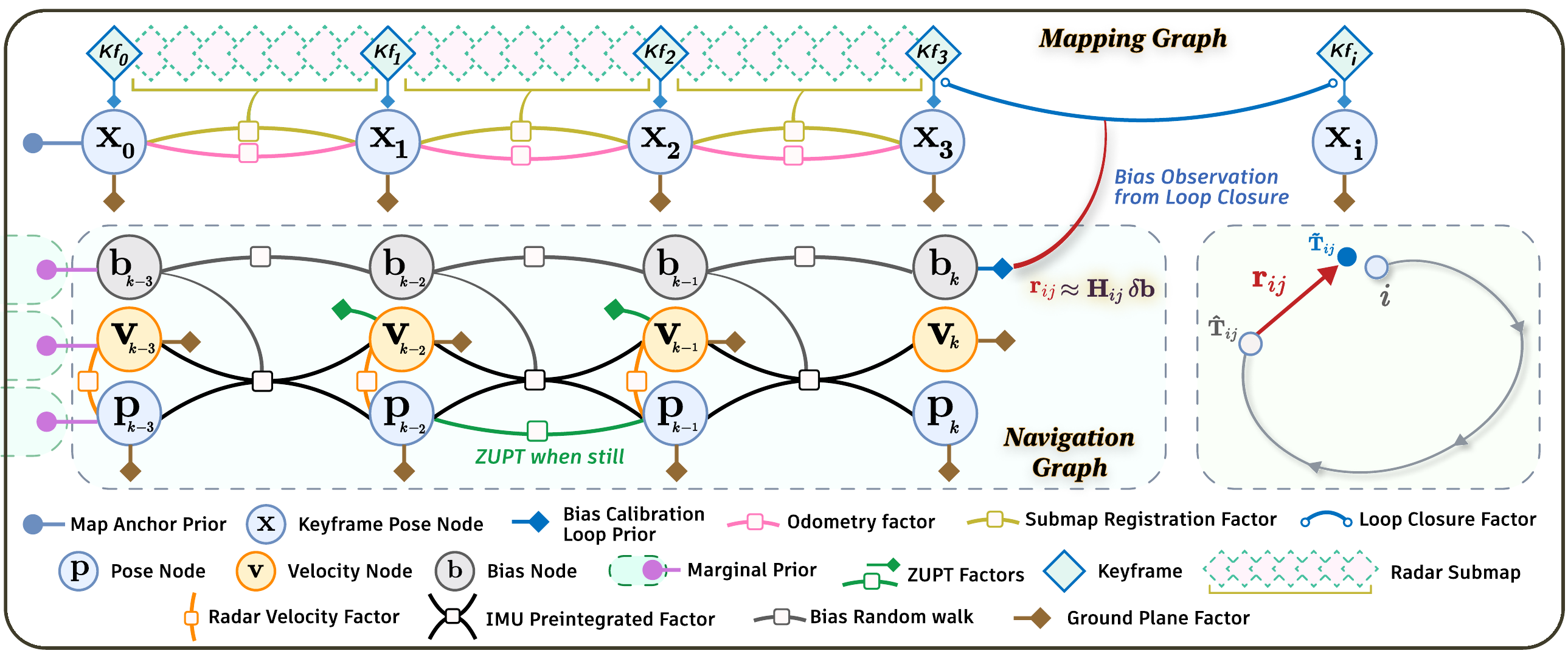}
    \caption{Hierarchical factor graph architecture of LoCal-RIO. (Top) The mapping graph maintains keyframe poses ($\mathbf{X}$) in the map frame, connected by odometry, submap registration, and robust loop closure factors. (Bottom) The navigation graph operates a fixed-lag window of pose ($\mathbf{p}$), velocity ($\mathbf{v}$), and bias ($\mathbf{b}$) nodes; states leaving the window are marginalized into a joint prior. (Right) Over a loop from keyframe $i$ to $j$, the relative pose estimated by the navigation graph, $\hat{\mathbf{T}}_{ij}$, deviates from the one measured by the loop closure, $\tilde{\mathbf{T}}_{ij}$. Their discrepancy $\mathbf{r}_{ij}$ is modeled as $\mathbf{r}_{ij} \approx \mathbf{H}_{ij}\,\delta\mathbf{b}$, where $\mathbf{H}_{ij}$ chains the preintegration bias Jacobians over the loop and $\delta\mathbf{b}$ is the bias correction, which enters the navigation graph as a bias calibration prior on $\mathbf{b}_k$.}
    \label{fig:hrio_fgo}
\end{figure*}

\section{Related Work}
\label{sec:related_work}

Radar-inertial odometry (RIO) tightly couples preintegrated IMU measurements with Doppler-derived ego-velocity, typically formulated as sliding-window factor graphs~\cite{23}, recursive filters~\cite{12}, or continuous-time Gaussian processes~\cite{4}. Operating over bounded temporal horizons, these estimators suffer from limited IMU bias observability.

\par Hierarchical SLAM typically pairs local odometry with global optimization to correct drift~\cite{1, 6, 18}. LIO-SAM~\cite{18} restricts its global backend to a pose graph, leaving IMU biases to be estimated within a local, resetting tracking window driven by mapping priors. Because this tracking graph directly integrates mapping poses, loop closures cause discontinuities in the control odometry, while bias estimates remain isolated from global corrections. In contrast, LoCal-RIO absorbs loop-closure corrections of position and heading in the map-to-odometry transform, and passes their systematic component to the tracking layer solely as an observation of the IMU bias.

\par To handle radar sparsity and noise, modern frontends leverage probabilistic models~\cite{2}, heteroscedastic covariance prediction~\cite{5}, and learning-based point matching~\cite{10, 17}. In contrast, LoCal-RIO introduces specular-filtered submapping, utilizing a maximum-SNR voxelization scheme to filter out low-power multipath clutter and preserve the raw physical coordinates of stable specular point targets. Calibration pipelines also jointly optimize spatial extrinsics and temporal offsets via tightly-coupled factor graphs~\cite{7, 8, 9}. While these methods enhance local registration, they operate independently of global optimization. LoCal-RIO is compatible with them, and additionally uses loop closures to refine the IMU calibration of the tracking layer.

\section{Methodology}

\subsection{System Overview}
We define the navigation state $\mathcal{X}_k$ at radar measurement time $t_k$ on the composite manifold $\mathcal{M} \triangleq \mathrm{SE}(3) \times \mathbb{R}^3 \times \mathbb{R}^6$:
\begin{equation}
    \mathcal{X}_k \triangleq \left\{ \mathbf{T}_k, \mathbf{v}_k, \mathbf{b}_k \right\} = \left\{ \mathbf{R}_k, \mathbf{p}_k, \mathbf{v}_k, \mathbf{b}_k^{\omega}, \mathbf{b}_k^a \right\}
\end{equation}
where $\mathbf{T}_k = (\mathbf{R}_k, \mathbf{p}_k) \in \mathrm{SE}(3)$ is the body pose in the odometry frame $O$, $\mathbf{v}_k \in \mathbb{R}^3$ the linear velocity in $O$, and $\mathbf{b}_k = [(\mathbf{b}_{k}^a)^\top, (\mathbf{b}_{k}^\omega)^\top]^\top \in \mathbb{R}^6$ the slowly-varying accelerometer and gyroscope biases. The mapping graph estimates keyframe poses $\mathbf{X}_i \in \mathrm{SE}(3)$ in the map frame $M$ and the transform $\mathbf{T}_{MO}$ (\cref{fig:hrio_fgo}).

Each graph computes the MAP estimate by minimizing the following nonlinear cost over the factor graph $\mathcal{G} = (\mathcal{V}, \mathcal{F})$ under zero-mean Gaussian noise:
\begin{equation}
    \mathcal{X}^* = \arg\min_{\mathcal{X}} \left( \| \mathbf{r}_{\text{prior}} \|_{\boldsymbol{\Sigma}_{0}}^2 + \sum_{k} \rho_k\bigl(\| \mathbf{r}_{k}(\mathcal{X}, \mathcal{Z}_k) \|_{\boldsymbol{\Sigma}_{k}}^2\bigr) \right),
\end{equation}
where $\rho_k$ is the identity unless a robust kernel is stated.

\subsection{Initialization and Extrinsic Alignment}
\label{sec:initialization}

\subsubsection{Static Gravity Alignment}
Let $\mathbf{a}_j \in \mathbb{R}^3$ be the specific force measured over a stationary period of $N = 200$ samples. The per-axis median $\bar{\mathbf{a}}$ defines the local vertical, and a rotation $\mathbf{R}_0 \in \mathrm{SO}(3)$ aligns $\bar{\mathbf{a}}$ with the gravity vector $\mathbf{g} = [0, 0, g]^\top$. This fixes roll and pitch uniquely. The yaw and position are unobservable and are set to zero, since they only define the origin of $O$; the initial velocity is taken from the first radar measurement.

\subsubsection{Extrinsic Tilt Refinement}
Roll and pitch errors of the IMU mounting would otherwise be absorbed by the accelerometer bias. Since one stationary period cannot separate them from the platform attitude, the tilt is estimated online from the mean specific-force direction over several quiet stretches, across which the platform attitude averages out. The estimate is adopted once it exceeds three times its standard error; the correction leaves yaw unchanged and is applied gradually.

\subsection{Ego-velocity Estimation}
\label{sec:ego_velocity}

The imaging radar provides the Doppler velocity of detected points. Assuming that the majority of points in a given frame belong to static objects, we can relate the Doppler radial velocity $v^d_n \in \mathbb{R}$ of point $n$, its 3D position in the sensor frame $\mathbf{p}_n \in \mathbb{R}^3$, and the sensor's linear ego-velocity $\mathbf{v}_R \in \mathbb{R}^3$ via:
\begin{equation}
    -v^d_n = \mathbf{v}_R^\top \frac{\mathbf{p}_n}{\|\mathbf{p}_n\|}
\end{equation}

To obtain an accurate sensor velocity resilient to dynamic outliers, we employ a robust least-squares approach. Let $N$ be the total number of radar points in the current scan, and let the Doppler measurement noise be modeled by a Gaussian distribution with variance $\sigma_d^2$. The optimal sensor velocity $\mathbf{v}_R^*$ is found by minimizing:
\begin{equation}
\label{eq:ceres_cost}
    \mathbf{v}_R^* = \arg\min_{\mathbf{v}_R} \sum_{n=1}^N \rho \left( \frac{1}{\sigma_d^2} \left\| \mathbf{v}_R^\top \frac{\mathbf{p}_n}{\|\mathbf{p}_n\|} + v^d_n \right\|^2 \right)
\end{equation}
where $\rho(\cdot)$ represents a robust error function (e.g., Cauchy or Welsch) to reject moving objects.

\subsubsection{Radar Velocity Factor}
The estimated ego-velocity $\mathbf{v}_R^*$, time-shifted by $-0.08$~s to compensate for the latency between the radar and IMU streams, provides a direct measurement of the sensor's linear velocity. Because the radar sensor is extrinsically offset from the IMU center by a rotation $\mathbf{R}_R^B$ and translation $\mathbf{p}_R^B$, the rotational motion of the vehicle induces a tangential lever-arm velocity.

First, the measured velocity is transformed to the Body frame $B$ and corrected using the angular rate $\hat{\boldsymbol{\omega}}_k$ and current gyroscope bias estimate $\mathbf{b}_k^{\omega}$:
\begin{equation}
    \mathbf{v}_{B}^{\text{meas}} = \mathbf{R}_R^B \mathbf{v}_R^* - \left( \hat{\boldsymbol{\omega}}_k - \mathbf{b}_k^{\omega} \right) \times \mathbf{p}_R^B
\end{equation}
The residual $\mathbf{r}_{\mathcal{D}}$ minimizes the difference between this corrected measurement and the estimated velocity $\mathbf{v}_k$, projected into the body frame via $\mathbf{R}_k^\top$:
\begin{equation}
    \mathbf{r}_{\mathcal{D}}(\mathcal{X}_k) = \mathbf{v}_{B}^{\text{meas}} - \mathbf{R}_k^\top \mathbf{v}_k
\end{equation}
It is weighted by $\boldsymbol{\Sigma}_{\mathcal{D}} = \mathrm{diag}(\sigma_{xy}^2, \sigma_{xy}^2, \sigma_{z}^2)$, with $\sigma_{xy} = 0.1$~m/s and $\sigma_z = 0.5$~m/s, and a Huber kernel with a threshold of $3\sigma$, which limits the influence of degenerate ego-velocity solutions.

\subsection{Navigation Graph}
\label{sec:navigation}

The navigation graph is an incremental fixed-lag smoother with one state per radar measurement. States older than $\tau_L = 3$~s are marginalized through the Schur complement into a dense prior, which preserves their information and cross-correlations. Scan registrations and loop closures do not enter this graph as pose constraints, and the newest estimate is published as odometry in $O$.

\subsubsection{IMU Preintegration}
Consecutive states $i$ and $j$ are connected by IMU preintegration factors. Preintegrated rotation, velocity, and position pseudo-measurements summarize the raw IMU stream between radar measurements. To avoid re-integration when the bias linearization point shifts during optimization, these terms are expressed through a first-order Taylor expansion around the current estimate $\bar{\mathbf{b}}_i$:
\begin{align}
\Delta\tilde{\mathbf{R}}_{ij}(\mathbf{b}_i) &\approx \Delta\tilde{\mathbf{R}}_{ij}(\bar{\mathbf{b}}_i) \, \mathrm{Exp}\bigl( \mathbf{J}_{\mathbf{b}^\omega}^{\Delta\mathbf{R}} \delta\mathbf{b}_i^\omega \bigr), \\[0.8ex]
\Delta\tilde{\mathbf{v}}_{ij}(\mathbf{b}_i) &\approx \Delta\tilde{\mathbf{v}}_{ij}(\bar{\mathbf{b}}_i) + \mathbf{J}_{\mathbf{b}^a}^{\Delta\mathbf{v}} \delta\mathbf{b}_i^a + \mathbf{J}_{\mathbf{b}^\omega}^{\Delta\mathbf{v}} \delta\mathbf{b}_i^\omega, \\[0.8ex]
\Delta\tilde{\mathbf{p}}_{ij}(\mathbf{b}_i) &\approx \Delta\tilde{\mathbf{p}}_{ij}(\bar{\mathbf{b}}_i) + \mathbf{J}_{\mathbf{b}^a}^{\Delta\mathbf{p}} \delta\mathbf{b}_i^a + \mathbf{J}_{\mathbf{b}^\omega}^{\Delta\mathbf{p}} \delta\mathbf{b}_i^\omega,
\end{align}
with $\delta\mathbf{b}_i = \mathbf{b}_i - \bar{\mathbf{b}}_i$ and Jacobians computed iteratively during preintegration. The residual compares the state prediction against the bias-corrected measurements:
\begin{multline}
\mathbf{r}_{\mathcal{I}}(\mathcal{X}_i, \mathcal{X}_j) = \\
\begin{bmatrix}
\mathrm{Log}\bigl( \Delta\tilde{\mathbf{R}}_{ij}(\mathbf{b}_i)^\top \mathbf{R}_i^\top \mathbf{R}_j \bigr) \\[1ex]
\mathbf{R}_i^\top \bigl( \mathbf{v}_j - \mathbf{v}_i - \mathbf{g} \Delta t_{ij} \bigr) - \Delta\tilde{\mathbf{v}}_{ij}(\mathbf{b}_i) \\[1ex]
\mathbf{R}_i^\top \bigl( \mathbf{p}_j - \mathbf{p}_i - \mathbf{v}_i \Delta t_{ij} - \frac{1}{2} \mathbf{g} \Delta t_{ij}^2 \bigr) - \Delta\tilde{\mathbf{p}}_{ij}(\mathbf{b}_i) \\[1ex]
\mathbf{b}_j - \mathbf{b}_i
\end{bmatrix},
\label{eq:imu_residual}
\end{multline}
weighted by the propagated covariance and the bias random walk. Within a short window this factor constrains relative bias changes effectively, but the gyroscope bias about the vertical axis remains weakly observable, which motivates the loop calibration of \cref{sec:loop_calibration}.

\subsubsection{Zero-Velocity Updates}

Stationary periods make absolute biases observable: when true velocity is zero, any non-zero integrated IMU motion must originate from bias errors. A stationary flag is raised when $\|\mathbf{v}_R^*\|$ and the bias-corrected angular rate remain below $0.05$~m/s and $0.05$~rad/s for $0.5$~s, and is released when either exceeds twice its threshold. Each state added while stationary receives a zero-velocity prior $\mathbf{v}_k = \mathbf{0}$ ($\sigma = 0.1$~m/s), and consecutive stationary states are connected by a zero-motion factor $\mathbf{T}_{k-1}^{-1}\mathbf{T}_k = \mathbf{I}$ ($\sigma = 0.5$). Combined with the IMU preintegration factor over the same interval, these constraints attribute the integrated motion to the bias states.

\subsubsection{Ground Plane Constraints}

Both graphs incorporate ground plane priors to constrain weakly observable degrees of freedom, particularly in open environments where vertical features are sparse. The plane is described by a unit normal $\mathbf{n}$ and offset $d$ in $O$. In the evaluated configuration, these parameters are fixed ($\mathbf{n} = \mathbf{e}_z$, $d = 0$); ground segmentation methods such as Patchwork++~\cite{patchwork} can provide them online. Let $\mathbf{R}_{\text{align}}$ be the rotation that aligns $\mathbf{n}$ with the $Z$-axis. For each state:
\begin{align}
\mathbf{e}_{\text{att}} &= \bigl[ \mathrm{Log}(\mathbf{R}_{\text{align}}^\top \mathbf{R}_k) \bigr]_{x,y}, \quad
\boldsymbol{\Sigma}_{\text{att}} = \sigma_{rp}^2 \mathbf{I}_2, \\[1.2ex]
\mathbf{e}_{\text{pos}} &= \bigl( \mathbf{R}_{\text{align}}^\top \mathbf{p}_k \bigr)_z - d, \quad
\boldsymbol{\Sigma}_{\text{pos}} = \sigma_z^2, \\[1.2ex]
\mathbf{e}_{\text{vel}} &= \bigl( \mathbf{R}_{\text{align}}^\top \mathbf{v}_k \bigr)_z, \quad
\boldsymbol{\Sigma}_{\text{vel}} = \sigma_{vz}^2,
\end{align}
constraining roll/pitch attitude, height above the ground plane, and normal velocity respectively. The mapping graph applies the attitude and height factors to each keyframe pose.

\subsection{Loop Closure as Bias Calibration}
\label{sec:loop_calibration}

A loop closure between keyframes $i$ and $j$ provides a measured relative pose $\tilde{\mathbf{T}}_{ij}$ with covariance $\tilde{\boldsymbol{\Sigma}}_{ij}$, while the navigation graph provides the estimate $\hat{\mathbf{T}}_{ij} = \hat{\mathbf{T}}_i^{-1}\hat{\mathbf{T}}_j$ from the poses recorded at both keyframes. In their discrepancy, the unobservable heading and position cancel, and the remainder is the navigation error accumulated between the two visits, whose systematic part is caused by the bias.

\subsubsection{Chained Bias Sensitivity}
Each keyframe segment $m$ stores its preintegrated terms $\Delta\mathbf{R}_m$, $\Delta\mathbf{v}_m$, $\Delta\mathbf{p}_m$, $\Delta t_m$ and bias Jacobians $\mathbf{D}^R_m$, $\mathbf{D}^v_m$, $\mathbf{D}^p_m \in \mathbb{R}^{3\times6}$. With the state at $i$ held fixed, the sensitivities of the relative rotation and position to a constant bias error are accumulated for $m = i{+}1, \dots, j$, starting from $\mathbf{A} = \mathbf{I}$ and $\mathbf{G} = \mathbf{H}^v = \mathbf{H}^p = \mathbf{0}$:
\begin{align}
\mathbf{H}^p &\leftarrow \mathbf{H}^p + \mathbf{H}^v \Delta t_m + \mathbf{A} \bigl( \mathbf{D}^p_m - [\Delta\mathbf{p}_m]_\times \mathbf{G} \bigr), \nonumber\\
\mathbf{H}^v &\leftarrow \mathbf{H}^v + \mathbf{A} \bigl( \mathbf{D}^v_m - [\Delta\mathbf{v}_m]_\times \mathbf{G} \bigr), \label{eq:chain}\\
\mathbf{G} &\leftarrow \Delta\mathbf{R}_m^\top \mathbf{G} + \mathbf{D}^R_m, \qquad \mathbf{A} \leftarrow \mathbf{A}\, \Delta\mathbf{R}_m, \nonumber
\end{align}
where the first two updates use $\mathbf{A}$ and $\mathbf{G}$ from the start of the segment. With the residual $\mathbf{r}_{ij} = [\mathrm{Log}( \hat{\mathbf{R}}_{ij}^\top \tilde{\mathbf{R}}_{ij})^\top,\, (\tilde{\mathbf{p}}_{ij} - \hat{\mathbf{p}}_{ij})^\top]^\top$ and the Jacobian $\mathbf{H}_{ij} = [(\mathbf{J}_r^{-1}\mathbf{G})^\top, (\mathbf{H}^p)^\top]^\top$, with the inverse right Jacobian $\mathbf{J}_r^{-1}$ of $\mathrm{SO}(3)$ at the rotational residual, the loop closure yields the linear observation $\mathbf{r}_{ij} \approx \mathbf{H}_{ij}\, \delta\mathbf{b} + \boldsymbol{\eta}$. Loops with an interrupted IMU record or an interval outside $[20, 600]$~s are not used.

\subsubsection{Noise Model}
In velocity-aided navigation, the velocity error remains bounded, whereas the attitude error accumulates and propagates into position through the travelled distance $L_{ij}$. The noise covariance is therefore $\boldsymbol{\Sigma}_{ij} = \tilde{\boldsymbol{\Sigma}}_{ij} + \boldsymbol{\Sigma}_{\text{rw}} + \kappa\, \boldsymbol{\Sigma}_{\text{w}}$. The bias random-walk term $\boldsymbol{\Sigma}_{\text{rw}}$ has the rotational block $\mathbf{G}\mathbf{Q}_b\mathbf{G}^\top T_{ij}$ over the interval duration $T_{ij}$ and a translational block proportional to $L_{ij}^2$; the white-noise term $\boldsymbol{\Sigma}_{\text{w}}$ contains the gyroscope angle random walk, its propagation through $L_{ij}$, and the integrated radar velocity noise. Loop closures spanning the same interval observe the same drift; $\kappa \geq 1$ is therefore one plus the fraction of $[t_i, t_j]$ already covered by previously used loop closures.

\subsubsection{Regression and Bias Prior}
Since $\mathbf{H}_{ij}$ is square and generically invertible, an unregularized fit would explain any residual. The fit is therefore regularized by the current bias marginal $\boldsymbol{\Sigma}_b$ of the navigation graph:
\begin{equation}
\delta\mathbf{b}^* = \arg\min_{\delta\mathbf{b}} \|\mathbf{r}_{ij} - \mathbf{H}_{ij}\delta\mathbf{b}\|^2_{\boldsymbol{\Sigma}_{ij}} + \|\delta\mathbf{b}\|^2_{\boldsymbol{\Sigma}_b}.
\end{equation}
The loop closure is accepted if the explained fraction $1 - \|\mathbf{r}_{ij} - \mathbf{H}_{ij}\delta\mathbf{b}^*\|^2_{\boldsymbol{\Sigma}_{ij}} / \|\mathbf{r}_{ij}\|^2_{\boldsymbol{\Sigma}_{ij}}$ is at least $0.25$. An accepted loop closure is added as a unary bias calibration prior on the next state:
\begin{equation}
\mathbf{e}_{\mathcal{L}}(\mathbf{b}_k) = \mathbf{r}_{ij} - \mathbf{H}_{ij}\bigl( \mathbf{b}_k - \hat{\mathbf{b}}_k \bigr), \qquad \boldsymbol{\Sigma}_{\mathcal{L}} = \boldsymbol{\Sigma}_{ij},
\label{eq:loop_factor}
\end{equation}
linearized at the current estimate $\hat{\mathbf{b}}_k$. The factor involves no pose; it changes only the bias used for subsequent integration.

\subsubsection{Deferred Use}
Since the bias update cannot be revoked, a loop closure is passed to the calibration only once its weight in the mapping graph is at least $0.1$ and its cycle-consistency statistic (\cref{sec:mapping_graph}) includes at least eight subsequent loop closures. Each loop closure is used at most once.

\subsection{Mapping Graph}
\label{sec:mapping_graph}

The mapping graph is a pose graph over all keyframe poses, without velocity or bias states. Its first node is anchored to the initial navigation pose, and it provides $\mathbf{T}_{MO} = \mathbf{X}_K \hat{\mathbf{T}}_K^{-1}$ at the latest keyframe $K$. A keyframe is created after $0.5$~m of translation, $0.2$~rad of rotation, or $0.5$~s ($10$~s while stationary).

\subsubsection{Odometry Factor}
Consecutive keyframes are connected by the relative pose $\hat{\mathbf{T}}_{i-1}^{-1}\hat{\mathbf{T}}_i$ of the navigation graph. Over the interval $\Delta t$, its roll and pitch variances follow the gyroscope angle random walk, since gravity bounds their error, whereas the unobservable yaw receives an additional random walk of $\sigma_\psi = 0.003$~rad$/\sqrt{\text{s}}$. The translational standard deviation grows by $5\%$ of the distance travelled, in addition to the integrated radar velocity noise. The same keyframes are also connected by a submap registration factor (\cref{sec:mapping}).

\subsubsection{Loop Closure Factors}
A loop closure is admitted if its keyframes are at least three keyframes and $5$~m of path apart. Its covariance is augmented by a per-axis noise floor estimated online from the loop innovations. The graph is solved with riSAM~\cite{mcgann_risam_2023}, an incremental solver with graduated non-convexity. Loop closures use the kernel $\rho(r, \mu) = \tfrac{1}{2} c^2 r^2 / (c^2 + r^{2\mu})$, where $r$ is the whitened residual, $c$ the $99\%$ quantile of $\chi^2_6$, and the convexity $\mu$ is annealed from $\mu_0 \in [0.86, 0.95]$. The lower bound limits the influence of an incorrect loop closure before evidence against it is available.

\subsubsection{Cycle Consistency}
Two loop closures $a$ and $b$ form a cycle with the odometry between their endpoints, $\mathbf{C}_{ab} = \tilde{\mathbf{T}}_a\, \hat{\mathbf{T}}_{a_1}^{-1}\hat{\mathbf{T}}_{b_1}\, \tilde{\mathbf{T}}_b^{-1}\, \hat{\mathbf{T}}_{b_0}^{-1}\hat{\mathbf{T}}_{a_0}$, which equals the identity for consistent measurements. Its normalized rotational error is $\chi_{ab} = \|\mathrm{Log}(\mathbf{R}(\mathbf{C}_{ab}))\|^2 / \bigl(3( 2\sigma_{c}^2 + \sigma_\psi^2 \Delta t_{ab})\bigr)$, where $\sigma_c$ is the measured loop rotation noise and $\Delta t_{ab}$ the duration of the two odometry segments. Pairs are stored for both loop closures, so earlier ones are also tested by later ones. Once loop closure $a$ has eight pairs, their median $\bar{\chi}_a$ sets $\mu_0 = \mathrm{clamp}(\bar{\chi}_a / 11.345,\, 0.86,\, 0.95)$; the statistics are recomputed every twenty updates, and loop closures whose $\mu_0$ increased are re-initialized. An incorrect loop closure is thus down-weighted once subsequent loop closures are available.

\begin{figure*}[t]
    \centering
    \includegraphics[width=0.98\linewidth]{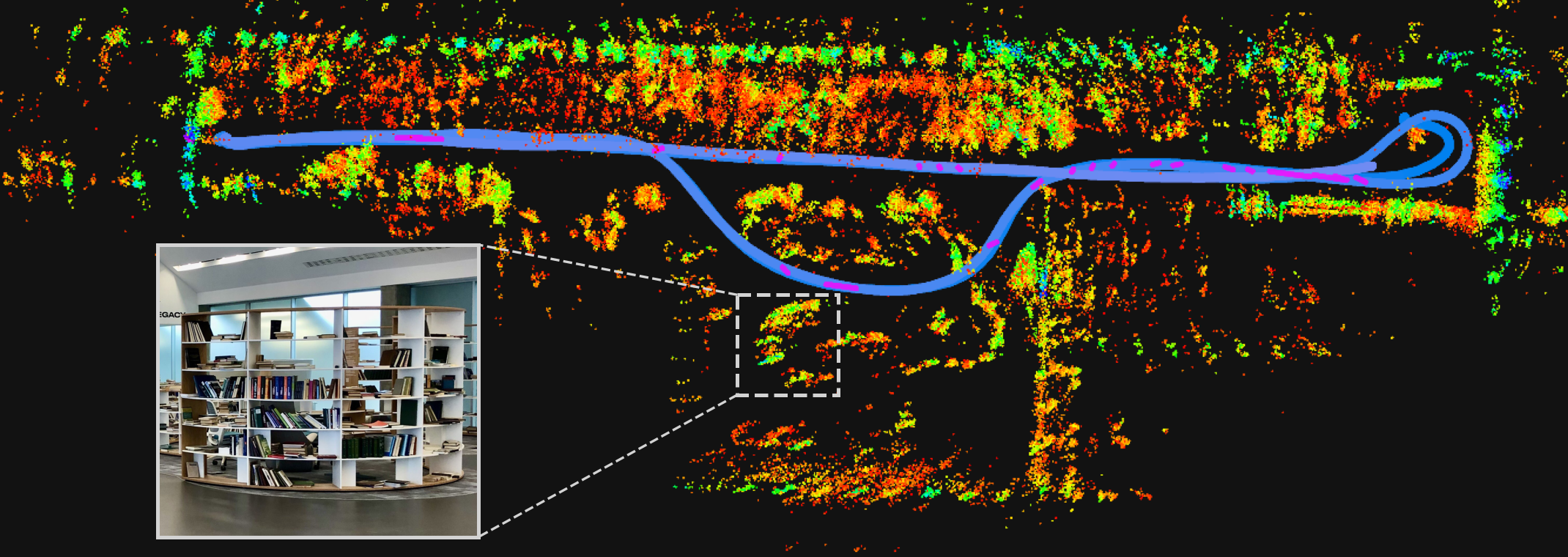}
    \caption{Top-down view of the radar map produced by LoCal-RIO for the \texttt{Library} sequence. Features with relatively low radar reflectivity, such as the wooden bookshelves, remain clearly distinguishable, and the overall spatial layout of the environment is preserved in the reconstructed map.}
    \label{fig:library_radar_map_bookshelf}
\end{figure*}

\subsection{Radar Mapping}
\label{sec:mapping}
The mapping module serves as a mid-tier estimator bridging raw scans and the mapping graph. Operating asynchronously as a local fixed-lag smoother on the published odometry, it isolates high-frequency tracking from mapping failures and scan-matching degeneracies.
\par \textit{1) Scan-to-Map Local Smoothing:} Each scan $\mathcal{S}_i$ at $t_i$ is optimized in an incremental smoother over a window $\tau = 2.5$~s via an odometry factor and a GICP factor against a local map of un-marginalized scans and nearby closed submaps.
\par \textit{2) Deferred Keyframe Assembly:} To prevent geometric blurring from un-converged estimates, submap assembly is deferred to the smoother's marginalization boundary. Upon marginalization (exiting $\tau$), scan poses are frozen at their converged values $\mathbf{T}_i^{\mathrm{frz}}$. Scans spanning keyframe $k$ are projected relative to the anchor pose $\mathbf{T}_k \triangleq \mathbf{T}_{a(k)}^{\mathrm{frz}}$ (the first scan in $k$), synthesizing an undistorted cloud:
\begin{equation}
    \mathcal{C}_k = \bigcup_{i \in k} \bigl(\mathbf{T}_k^{-1}\,\mathbf{T}_i^{\mathrm{frz}}\bigr) \cdot \mathcal{S}_i.
\label{eq:keyframe_cloud}
\end{equation}
\par \textit{3) Specular-Voxelized Submaps:} Unlike centroid averaging, which blends specular returns with multipath clutter, the clouds for the published map and place recognition keep in each voxel $\mathcal{G}_b$ the raw coordinates of the peak specular scatterer, $\mathbf{p}_b = \arg\max_{\mathbf{p} \in \mathcal{G}_b} \mathrm{SNR}(\mathbf{p})$, discarding low-power noise.
\par \textit{4) Sequential Alignment:} GICP registers $\mathcal{C}_k$ against the union $\mathcal{M}_k$ of the three most recently closed submaps, using the smoother-estimated relative pose $\mathbf{T}_{k-1}^{-1}\mathbf{T}_k$ as the initial guess, yielding refined transform $\Delta\hat{\mathbf{T}}_{k-1,k}$ and information matrix $\boldsymbol{\Lambda}_{\mathrm{gicp}}$. If validated against convergence, fitness, inlier, and deviation thresholds, it is added to the mapping graph as a submap registration factor with a Huber kernel, covariance $\boldsymbol{\Sigma}_{\mathrm{map}} = \boldsymbol{\Lambda}_{\mathrm{gicp}}^{-1}$, and error vector:
\begin{equation}
    \mathbf{e}_{\mathrm{map}} = \mathrm{Log} \bigl( \Delta\hat{\mathbf{T}}_{k-1,k}^{-1}\;\mathbf{X}_{k-1}^{-1}\,\mathbf{X}_k \bigr),
\label{eq:map_factor}
\end{equation}
with the smoother estimate serving as a fallback during GICP failures.
\par \textit{5) Place Recognition:} Each closed submap is merged with up to six neighboring submaps within $30$~m, so that the query describes the same neighborhood regardless of the direction of travel. A candidate becomes a loop closure only after its relative pose has been estimated and verified.

\begin{figure}[H]
    \centering
    \includegraphics[width=0.8\linewidth]{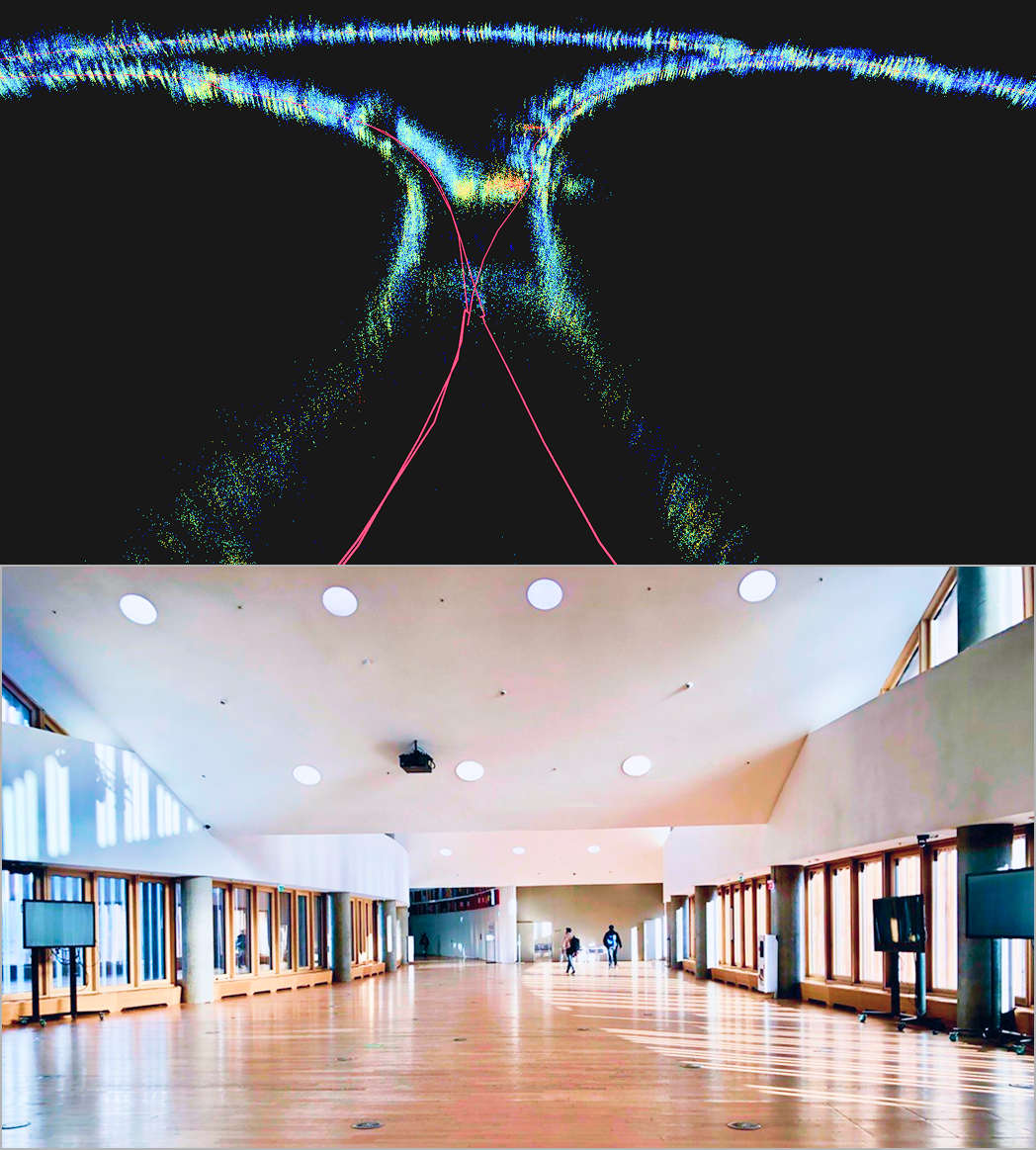}
    \caption{The concave hall with its curved walls remains geometrically consistent in the map throughout the motion, retaining the fidelity of the environment.}
    \label{fig:main_hall_image_map}
\end{figure}

\begin{figure*}[t]
    \vspace*{-1em}
    \centering
    \includegraphics[width=\textwidth]{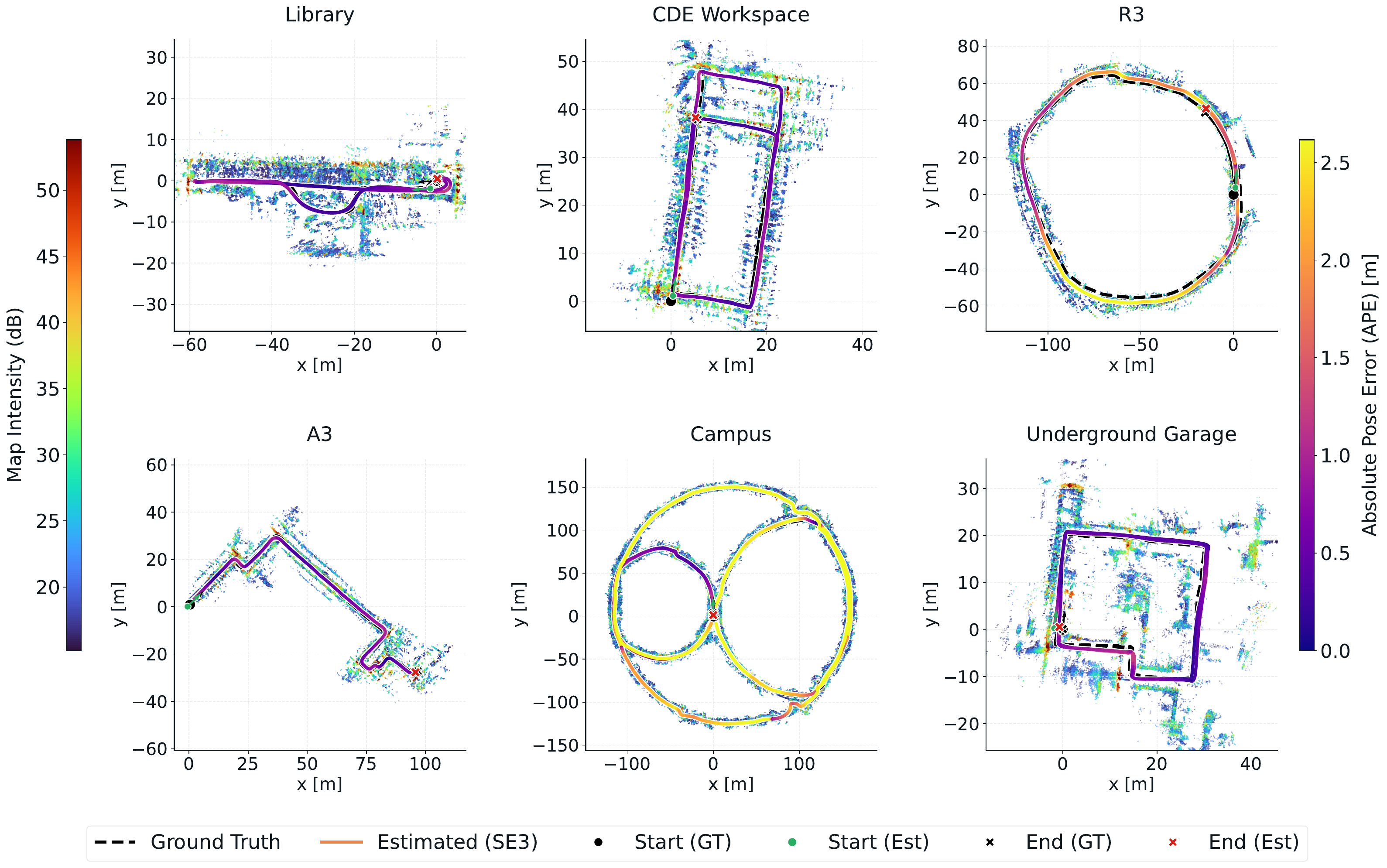}
    \caption{Qualitative evaluation across various sequences with different indoor environment characteristics: the Library sequence features a long, dynamic environment with wood and furniture; the CDE workspace features a rectangular corridor around a meeting room with glass walls; and the Underground garage sequence is an empty underground area with mostly cement corridors and a firefighting garage.}
    \label{fig:trajectories_map_evaluations}
\end{figure*}

\begin{figure}[H]
    \centering
    \includegraphics[width=0.5\linewidth]{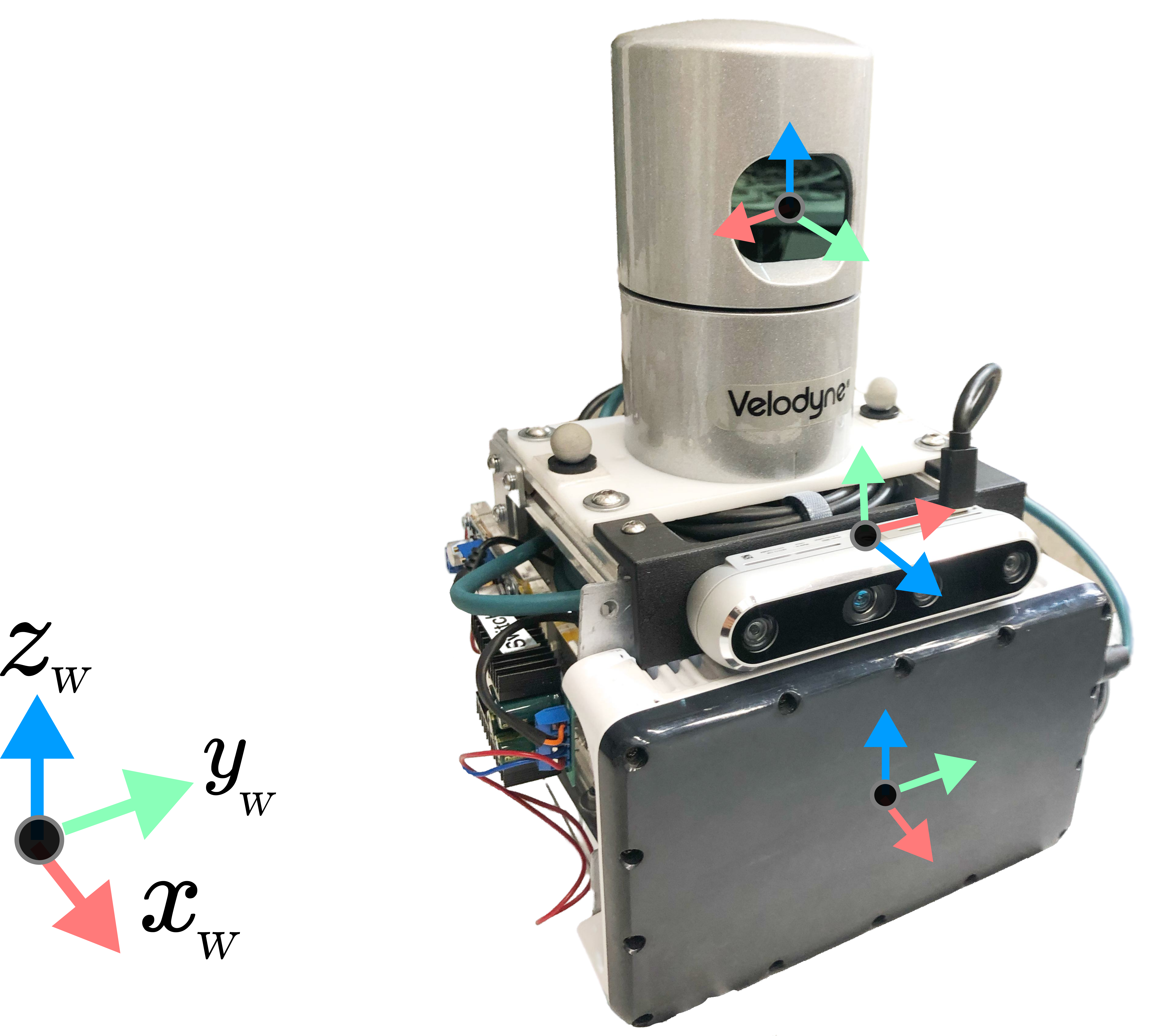}
    \caption{Platform setup used in this work. It features an Imaging FMCW radar (Integrant HD-Radar), which outputs 4D point clouds at a rate of 20\,Hz, and a commercial-grade Bosch BMI085 IMU (built into a RealSense D455 camera), with an Intel NUC7i7BNH as the compute unit.}
    \label{fig:sensor_setup}
\end{figure}

\section{Experiments}
LoCal-RIO is implemented in C++ using GTSAM\cite{gtsam}. The navigation graph and the mapping smoother use incremental fixed-lag smoothers based on the ISAM2 solver\cite{isam2}, and the mapping graph is solved with riSAM~\cite{mcgann_risam_2023}. Ceres solver\cite{Agarwal_Ceres_Solver_2022} was used for the compact nonlinear least squares problem of the instantaneous ego velocity estimation. The setup used for data collection and experiments is shown in \Cref{fig:sensor_setup}.


\begin{table}[H]    
\centering
    \begin{threeparttable}
    \caption{Quantitative Evaluation}
    \label{tab:hrio_complete}
    \begin{tabular}{@{} l c c c @{}}
        \toprule
        \textbf{Sequence} & \textbf{ATE [m]} & \multicolumn{2}{c}{\textbf{RPE}} \\
        \cmidrule(lr){3-4}
        \footnotesize\textit{name (distance, speed)} & & \textbf{[m]} & \textbf{[deg]} \\
        \midrule
        \footnotesize\textit{A3 (158.8 m, 1.13 m/s)} & 0.710 & 0.085 & 13.148 \\
        \footnotesize\textit{Underground Garage (238.2 m, 1.06 m/s)} & 0.766 & 1.430 & 6.304 \\
        \footnotesize\textit{Library (252.9 m, 1.27 m/s)} & 0.693 & 1.327 & 5.714 \\
        \footnotesize\textit{CDE Workspace (276.2 m, 0.65 m/s)} & 0.748 & 1.395 & 2.889 \\
        \footnotesize\textit{R3 (421.5 m, 1.57 m/s)} & 1.935 & 1.388 & 0.775 \\
        \footnotesize\textit{Campus (2203.5 m, 1.16 m/s)} & 3.440 & 0.149 & 4.695 \\
        \bottomrule
        \end{tabular}
        \begin{tablenotes}
            \item[]\footnotesize RPE is computed over all 1\,m relative pose pairs.
        \end{tablenotes}
    \end{threeparttable}
\end{table}

\subsection{Reference Trajectory Acquisition}
We adopt the localization methodology of \cite{localize_on_priors}, which augments FAST-LIO2 \cite{fast_lio2} with a degeneration-aware prior map factor for robust graph-based pose estimation. Dense prior maps are acquired with a Livox Mid-360, whose non-repetitive scanning pattern densifies the map under slow motion, and the platform carries a Velodyne HDL-32E; this pipeline provides the pseudo-ground truth.  We evaluate the trajectories according to the standard metrics using \cite{evo}.

\begin{table*}[htbp]
    \centering
    \begin{threeparttable}
        \caption{Impact of Factor Ablation on the Odometry Accuracy of the Navigation Graph}
        \label{tab:hrio_ablation}
        \begin{tabular}{@{} l *{12}{c} @{}}
            \toprule
            & \multicolumn{3}{c}{\textbf{Full Odometry}} & \multicolumn{3}{c}{\textbf{w/o Loop Calibration}} & \multicolumn{3}{c}{\textbf{w/o ZUPT}} & \multicolumn{3}{c}{\textbf{w/o GICP}} \\
            \cmidrule(lr){2-4} \cmidrule(lr){5-7} \cmidrule(lr){8-10} \cmidrule(lr){11-13}
            \textbf{Sequence} & \textbf{ATE} & \multicolumn{2}{c}{\textbf{RPE}} & \textbf{ATE} & \multicolumn{2}{c}{\textbf{RPE}} & \textbf{ATE} & \multicolumn{2}{c}{\textbf{RPE}} & \textbf{ATE} & \multicolumn{2}{c}{\textbf{RPE}} \\
            & \textbf{[m]} & \textbf{[m]} & \textbf{[deg]} & \textbf{[m]} & \textbf{[m]} & \textbf{[deg]} & \textbf{[m]} & \textbf{[m]} & \textbf{[deg]} & \textbf{[m]} & \textbf{[m]} & \textbf{[deg]} \\
            \midrule
            \textit{A3} & 0.737 & 0.083 & 4.464 & 1.056 & 0.156 & 9.317 & 1.025 & 0.170 & 5.603 & 1.059 & 0.163 & 7.212 \\
            \textit{Underground Garage} & 5.036 & 0.385 & 7.954 & 6.339 & 0.313 & 4.960 & 4.639 & 0.355 & 5.341 & 7.913 & 0.375 & 9.738 \\
            \textit{Library} & 1.423 & 0.257 & 6.496 & 5.856 & 0.237 & 5.133 & 3.927 & 0.283 & 7.121 & 4.573 & 0.257 & 6.665 \\
            \textit{CDE Workspace} & 1.985 & 0.180 & 3.774 & 2.752 & 0.300 & 9.624 & 2.240 & 0.204 & 4.194 & 5.116 & 0.284 & 4.315 \\
            \textit{R3} & 8.618 & 0.112 & 1.115 & 8.814 & 0.114 & 1.244 & 8.494 & 0.109 & 1.280 & 4.126 & 0.101 & 1.070 \\
            \midrule
            \textbf{Avg. Degradation} & & & & \textbf{+84.3\%} & \textbf{+25.9\%} & \textbf{+43.3\%} & \textbf{+43.7\%} & \textbf{+23.3\%} & \textbf{+5.6\%} & \textbf{+85.6\%} & \textbf{+28.4\%} & \textbf{+19.4\%} \\
            \bottomrule
        \end{tabular}
        \begin{tablenotes}
            \item \footnotesize RPE is computed over all 1\,m relative pose pairs.
        \end{tablenotes}
    \end{threeparttable}
\end{table*}

\begin{table}[!t]
    \centering
    \caption{Graph Optimization Runtime Analysis}
    \label{tab:graph_optimization_runtime}
    \begin{tabular}{@{} l c c c c c @{}}
        \toprule
        \textbf{Graph Size} & \textbf{N} & \textbf{Avg.} & \textbf{Std. Dev.} & \textbf{Min} & \textbf{Max} \\
        & & \textbf{[ms]} & \textbf{[ms]} & \textbf{[ms]} & \textbf{[ms]} \\
        \midrule
        \multicolumn{6}{c}{\textit{Mapping Graph}} \\
        \midrule
        \hspace{0.5em} Small & 2637 & 8.42 & 8.11 & 0.22 & 62.67 \\
        \hspace{0.5em} Medium & 6002 & 42.49 & 41.72 & 4.10 & 294.80 \\
        \hspace{0.5em} Large & 4343 & 254.81 & 350.44 & 24.69 & 2298.83 \\
        \midrule
        \multicolumn{6}{c}{\textit{Navigation Graph}} \\
        \midrule
        \hspace{0.5em} Per navigation step & 28636 & 6.34 & 3.81 & 0.72 & 269.67 \\
        \bottomrule
    \end{tabular}
\end{table}

\begin{figure*}[t]
    \includegraphics[width=0.98\textwidth]{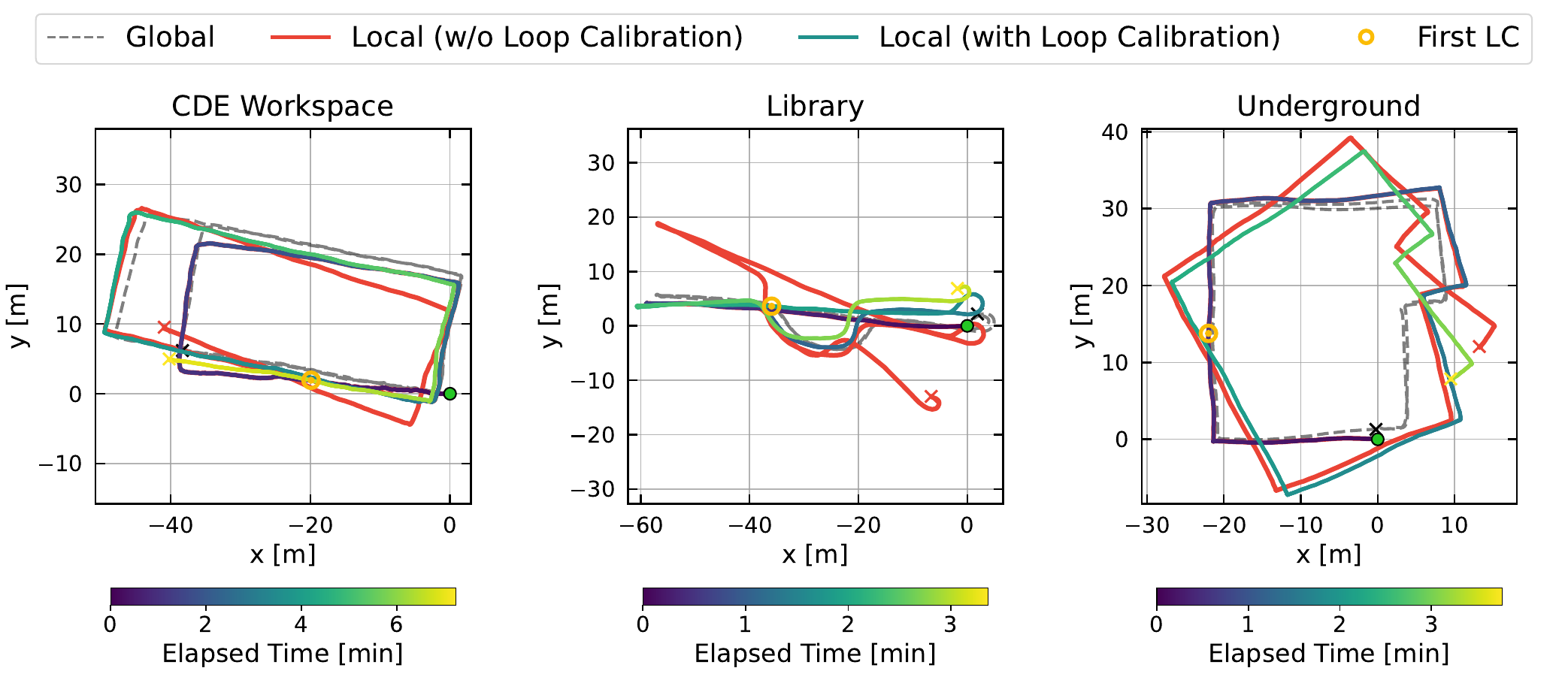}
    \caption{\textbf{Loop Calibration Ablation}: Trajectory comparison in three evaluation environments; \textit{Global} and \textit{Local} denote the mapping graph and navigation graph estimates. Without loop calibration, the unconstrained yaw gyroscope bias (\cref{fig:gyro_z_bias}) causes heading drift and trajectory divergence, most visibly in the \textit{Library} sequence. With loop calibration, the loop closures of the mapping graph constrain the bias, and the odometry remains consistent across all runs.}
    \label{fig:bias_ablation_trajectories_plot}
\end{figure*}

\subsection{Ablation Study}
We disable key components individually and measure the resulting impact. Disabling loop calibration confines the biases to the short-term observability of the navigation graph, and the uncorrected bias drift leads to severe accuracy degradation. Disabling ZUPT removes the explicit stationary constraints; since near-zero radar velocities still constrain a stationary platform, this ablation measures the benefit of the explicit constraints, which reduce drift even in sequences with few stops. Disabling GICP registration factors also degrades accuracy, indicating that the framework relies on these constraints. Quantitative findings for all ablations are reported in \Cref{tab:hrio_ablation}.

\begin{figure}[H]
    \includegraphics[width=\linewidth]{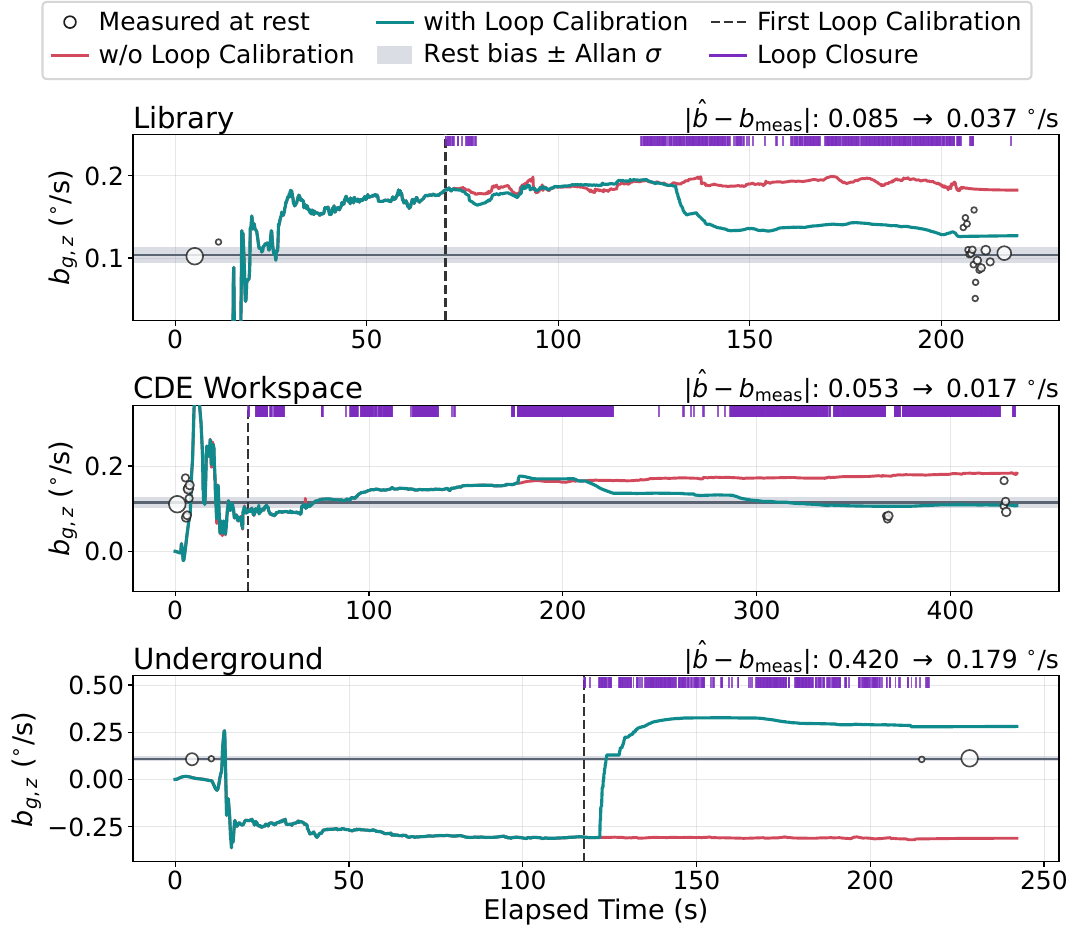}
    \caption{Gyroscope $z$-axis bias with and without loop calibration, on three sequences. Circles mark the bias measured directly as the mean gyroscope reading over stationary stretches---ground truth for this quantity, obtained without the estimator---with within-stretch standard errors; stretches occur when the platform starts and stops, so they cluster at the ends of each run. The shaded strip is the Allan bias instability, i.e., the expected variation of the bias about the measured mean. Loop calibration brings the estimate $2.8\times$, $1.5\times$, and $1.9\times$ closer to the measured bias (medians, top right of each panel). The step at the first calibration on \textit{Underground} is expected: a loop closure affects the inertial state only through a factor on the bias, never as a pose constraint, so the correction is not propagated into the published trajectory. }
    \label{fig:gyro_z_bias}
\end{figure}

\subsection{Runtime Analysis}
We measure the execution time of the complete optimization step of each graph and examine how the number of variables affects latency. All mapping graphs used in this evaluation include loop closures.

\section{Limitations}
The bias calibration assumes a constant bias over the loop interval and discards the pose--bias cross-covariance. Since a bias update cannot be revoked, robustness to false loop closures relies on the consistency test. The yaw of the IMU mounting must be calibrated beforehand.

\section{Conclusion}
LoCal-RIO separates low-latency tracking in a fixed-lag navigation graph from a keyframe mapping graph that absorbs the unobservable heading and position. Loop closures supported by consistent peers calibrate the IMU biases through a prior that constrains no pose, reducing drift at its source while the odometry remains smooth. Experiments confirm the accuracy and efficiency of the approach in visually degraded environments.


\bibliographystyle{IEEEtran}
\bibliography{references}

\end{document}